\documentclass[10pt,twocolumn]{article}

\usepackage{times}
\usepackage{amsmath,amssymb,amsfonts}
\usepackage{graphicx}
\usepackage{booktabs}
\usepackage{multirow}
\usepackage{enumitem}
\usepackage{xcolor}
\usepackage{hyperref}
\usepackage{amsthm}
\usepackage{mathrsfs}
\usepackage{tikz}
\usetikzlibrary{arrows.meta, positioning, fit, backgrounds, calc}
\usepackage[
    top=0.65in,
    bottom=0.75in,
    left=0.68in,
    right=0.68in,
    columnsep=0.25in
]{geometry}

\usepackage{titling}
\pretitle{\begin{center}\Large\bfseries}
\posttitle{\par\end{center}\vspace{-0.8em}}

\preauthor{\begin{center}\normalsize}
\postauthor{\par\end{center}\vspace{-1.2em}}

\predate{\begin{center}\normalsize}
\postdate{\par\end{center}\vspace{-2.0em}}

\newtheorem{definition}{Definition}

\newcommand{\paperTitle}{Bounded Behavioral Indistinguishability for Black-Box LLM Distillation}

\title{\paperTitle}

\date{}

\author{
\Large {Munawar Hasan} \vspace{0.3em} \\
\large {Michigan Technological University, Houghton, MI, USA}\\
\large \texttt{munawarh@mtu.edu}
}

\begin{document}

\maketitle

\begin{abstract}
Black-box LLM distillation is usually evaluated as an output-matching problem: a student is considered successful when its responses are semantically similar to, or task-consistent with, those of a teacher. However, output similarity does not imply that the student is behaviorally indistinguishable from the model it imitates. We introduce \textit{bounded behavioral indistinguishability}, formalized as $(\epsilon,q,t,\mathbb{A})$-behavioral indistinguishability over an explicit prompt distribution, where $\epsilon$ bounds distinguishing advantage, $q$ bounds oracle queries, $t$ bounds computation, and $\mathbb{A}$ denotes the adversary class.

We instantiate this notion on Qwen and Llama teacher--student pairs using a controlled $5,000$-prompt behavioral probe suite. For each family, we compare the teacher with both the base student and the LoRA-distilled student, measuring whether distillation reduces distinguishability rather than merely improving similarity. LoRA raises semantic similarity from $0.788$ to $0.862$ for Qwen and from $0.814$ to $0.874$ for Llama. Yet adversarial evaluation reveals remaining behavioral differences: learned discriminators retain nonzero advantage, and pairwise category analysis shows artifacts concentrated in style/format, robustness, and domain-technical prompts. A pairwise teacher-identification adversary confirms this trend. With a different-family Llama judge and A/B-swap consistency filtering, Qwen distinguishing advantage drops from $0.158$ for the base student to $0.081$ after LoRA distillation. Query-budget experiments show that disagreement-guided acquisition does not consistently outperform stratified random sampling, indicating that coverage and diversity remain strong baselines. Our results show that semantic fidelity is useful but insufficient: black-box LLM distillation requires bounded, adversarial, and category-aware evaluation.
\end{abstract}

\section{Introduction}
\label{sec:introduction}
Large language models (LLMs) are increasingly accessed through black-box APIs: users submit prompts and observe generated responses, while model weights, gradients, training data, and internal activations remain hidden. In many cases, even token-level probability information is unavailable or only partially exposed~\cite{zhu2025auditing, zhang2025blackboxopt, liu2026beyond}. This interface creates two closely related settings. In benign settings, organizations may distill larger models into smaller students to reduce latency, lower inference cost, enable on-device deployment, or adapt behavior to a specific domain. In adversarial settings, the same prompt-response interface can be used for model extraction, where an external party trains a surrogate model to imitate a deployed teacher. In both cases, the core question is the same: how closely does the student's observable behavior match the teacher's?

Black-box LLM distillation is usually evaluated as an output-matching problem: a student is considered successful when its responses are semantically similar to, task-consistent with, or lexically close to those of the teacher. These metrics are useful, but they do not answer a more operational question:

\begin{quote}
\emph{Can an adversary tell whether a response was generated by the teacher or by the distilled student?}
\end{quote}

This distinction matters because output similarity can hide behavioral differences. A medical assistant may preserve the main answer while omitting cautionary language; a coding assistant may reproduce the algorithm while revealing systematic formatting artifacts; an enterprise chatbot may match factual content while differing in refusal behavior, privacy warnings, or instruction-conflict resolution. In such cases, semantic similarity may overstate behavioral transfer.

We introduce \textit{bounded behavioral indistinguishability} for black-box LLM distillation. A student $S$ is $(\epsilon,q,t,\mathbb{A})$-behavioral indistinguishable from a teacher \(T\) if no adversary in a specified class $\mathbb{A}$, operating with query budget \(q\) and computational budget $t$, can distinguish teacher outputs from student outputs with advantage greater than $\epsilon$ over an explicit prompt distribution. This makes indistinguishability a bounded empirical claim rather than an absolute property of the model: behavioral transfer is measured by the residual advantage of bounded adversaries, not by output similarity alone.

This perspective yields an evaluation methodology based on empirical distinguishing advantage. Instead of relying on a single similarity metric, we evaluate teacher--student pairs using complementary adversarial and behavioral tests: learned discriminators, semantic similarity, category-wise probes, policy-level agreement, and pairwise teacher-identification judges. Each test captures a different residual signal. Response-only discriminators may detect style or length artifacts; prompt-response discriminators test whether responses are teacher-like in context; policy-level evaluators focus on safety-relevant behavior; and pairwise judges ask whether, given two responses to the same prompt, the teacher-generated output can be identified.

We instantiate this approach on Qwen~\cite{qwen} and Llama~\cite{llama} teacher--student pairs. For each family, we compare the teacher against both the original base student and a LoRA-distilled student trained on teacher-generated black-box responses using Low-Rank Adaptation~\cite{hu2022lora}. Our controlled 5,000-prompt behavioral probe suite covers general question answering, reasoning, coding, summarization, style and format control, ambiguous prompts, safety-boundary prompts, instruction conflict, domain-technical questions, and robustness perturbations. The suite is not intended to model a natural deployment distribution; rather, it provides a controlled probe distribution for measuring where behavioral imitation succeeds and where distinguishable artifacts remain.

Recent work has cautioned that indistinguishability-style measurements should be interpreted within a specified threat model, rather than treated as universal guarantees~\cite{liu2026beyond}. Our results support this bounded view: LoRA distillation improves semantic similarity across both Qwen and Llama, and learned discriminators become less effective at separating teacher outputs from student outputs, but residual distinguishability remains.

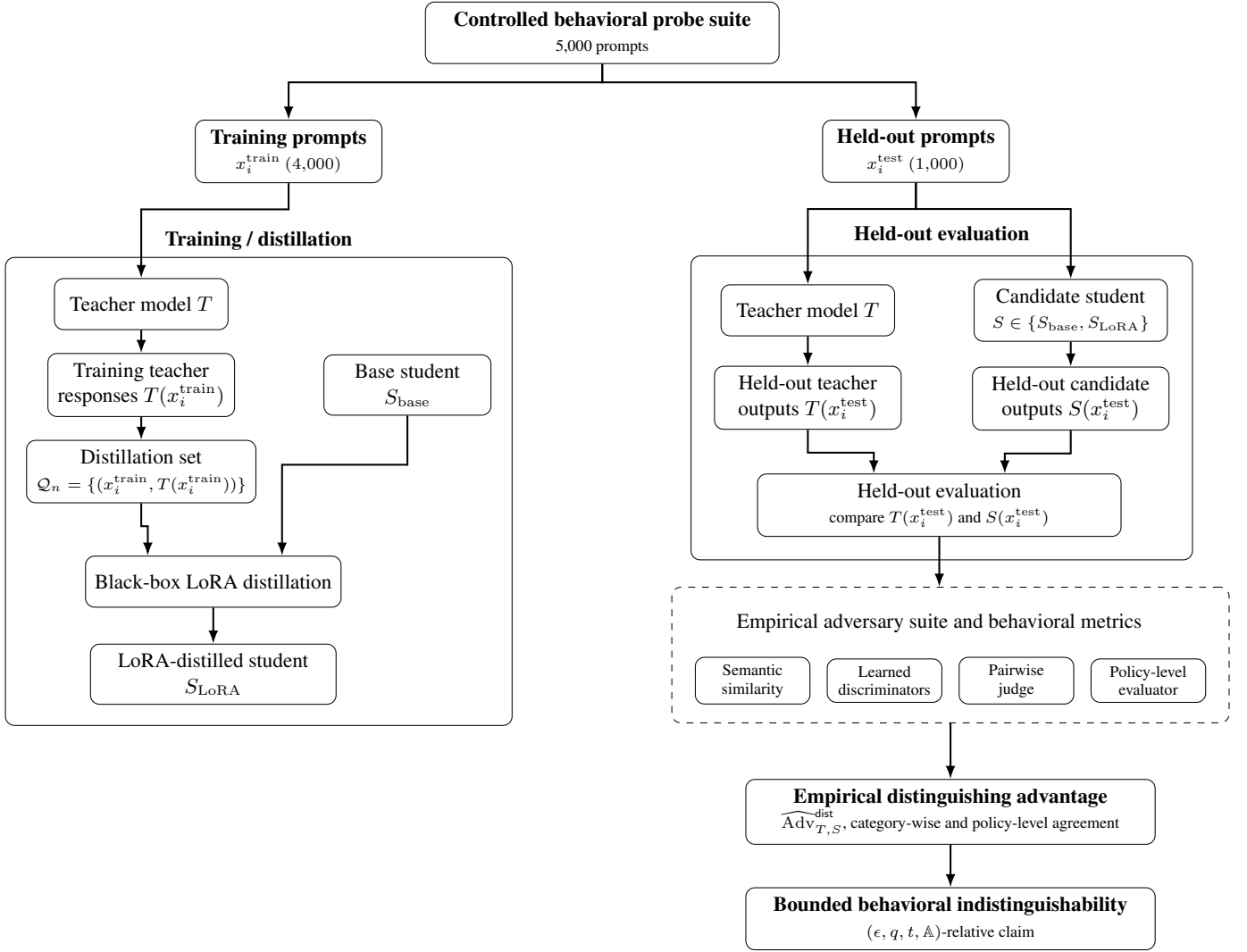
\begin{figure*}[t]
\centering
\resizebox{\textwidth}{!}{%
\begin{tikzpicture}[
    node distance=0.65cm and 0.75cm,
    box/.style={
        draw,
        rounded corners,
        align=center,
        minimum height=0.78cm,
        font=\small,
        inner sep=4pt
    },
    widebox/.style={
        draw,
        rounded corners,
        align=center,
        minimum height=0.82cm,
        font=\small,
        inner sep=4pt
    },
    titlebox/.style={
        draw,
        rounded corners,
        align=center,
        minimum width=5.4cm,
        minimum height=0.86cm,
        font=\small,
        inner sep=4pt
    },
    groupbox/.style={
        draw,
        rounded corners,
        inner sep=9pt
    },
    metricbox/.style={
        draw,
        rounded corners,
        align=center,
        minimum width=1.75cm,
        minimum height=0.65cm,
        font=\scriptsize,
        inner sep=3pt
    },
    arrow/.style={-{Latex[length=2.0mm]}, thick},
    line/.style={thick}
]

\node[titlebox] (suite) at (0,0)
{\textbf{Controlled behavioral probe suite}\\
{\scriptsize 5,000 prompts}};

\node[widebox, minimum width=2.85cm, below left=0.85cm and 0.65cm of suite] (trainprompts)
{\textbf{Training prompts}\\
{\scriptsize \(x_i^{\mathrm{train}}\) \((4{,}000)\)}};

\node[widebox, minimum width=2.85cm, below right=0.85cm and 0.65cm of suite] (testprompts)
{\textbf{Held-out prompts}\\
{\scriptsize \(x_i^{\mathrm{test}}\) \((1{,}000)\)}};

\draw[arrow] (suite.south) -- ++(0,-0.28) -| (trainprompts.north);
\draw[arrow] (suite.south) -- ++(0,-0.28) -| (testprompts.north);

\node[box, minimum width=2.65cm, below=1.45cm of trainprompts, xshift=-2.25cm] (trainteacher)
{Teacher model \(T\)};

\node[box, minimum width=2.85cm, below=0.35cm of trainteacher] (trainresp)
{Training teacher\\
responses \(T(x_i^{\mathrm{train}})\)};

\node[box, minimum width=3.25cm, below=0.35cm of trainresp] (distset)
{Distillation set\\
{\scriptsize \(\mathcal{Q}_n=\{(x_i^{\mathrm{train}},T(x_i^{\mathrm{train}}))\}\)}};

\node[box, minimum width=2.55cm, right=1.35cm of trainresp] (basestudent)
{Base student\\
\(S_{\mathrm{base}}\)};

\node[box, minimum width=3.75cm, below=0.80cm of distset, xshift=1.10cm] (lora)
{Black-box LoRA distillation};

\node[box, minimum width=3.75cm, below=0.55cm of lora] (lorastudent)
{LoRA-distilled student\\
\(S_{\mathrm{LoRA}}\)};

\draw[arrow] (trainprompts.south) -- ++(0,-0.40) -| (trainteacher.north);
\draw[arrow] (trainteacher) -- (trainresp);
\draw[arrow] (trainresp) -- (distset);

\draw[arrow]
    (distset.south) -- ++(0,-0.35) -| ([xshift=-1.cm]lora.north);

\draw[arrow]
    (basestudent.south) -- ++(0,-0.75) -| ([xshift=1.05cm]lora.north);

\draw[arrow] (lora) -- (lorastudent);

\begin{pgfonlayer}{background}
\node[
    groupbox,
    fit=(trainteacher)(trainresp)(distset)(basestudent)(lora)(lorastudent),
    label={[font=\small\bfseries, yshift=-0.01cm]above:Training / distillation}
] (traingroup) {};
\end{pgfonlayer}

\node[box, minimum width=2.65cm, below=1.55cm of testprompts, xshift=-1.65cm] (evalteacher)
{Teacher model \(T\)};

\node[box, minimum width=2.85cm, below=0.45cm of evalteacher] (testteacher)
{Held-out teacher\\
outputs \(T(x_i^{\mathrm{test}})\)};

\node[box, minimum width=2.95cm, right=1.20cm of evalteacher] (candidate)
{Candidate student\\
{\scriptsize \(S \in \{S_{\mathrm{base}},S_{\mathrm{LoRA}}\}\)}};

\node[box, minimum width=3.00cm, below=0.38cm of candidate] (teststudent)
{Held-out candidate\\
outputs \(S(x_i^{\mathrm{test}})\)};

\node[box, minimum width=5.55cm, below=1.15cm of $(testteacher)!0.5!(teststudent)$] (heldouteval)
{Held-out evaluation\\
{\scriptsize compare \(T(x_i^{\mathrm{test}})\) and \(S(x_i^{\mathrm{test}})\)}};

\draw[arrow] (testprompts.south) -- ++(0,-0.40) -| (evalteacher.north);
\draw[arrow] (testprompts.south) -- ++(0,-0.40) -| (candidate.north);
\draw[arrow] (evalteacher) -- (testteacher);
\draw[arrow] (candidate) -- (teststudent);

\draw[arrow] (testteacher.south) -- ++(0,-0.35) -| ([xshift=-1.0cm]heldouteval.north);
\draw[arrow] (teststudent.south) -- ++(0,-0.35) -| ([xshift=1.0cm]heldouteval.north);

\begin{pgfonlayer}{background}
\node[
    groupbox,
    fit=(evalteacher)(candidate)(testteacher)(teststudent)(heldouteval),
    inner sep=10pt
] (evalgroup) {};
\end{pgfonlayer}

\node[font=\small\bfseries, above=0.10cm of evalgroup.north] (evalgrouplabel)
{Held-out evaluation};

\node[font=\small] (metrictitle) 
    [below=1.05cm of heldouteval]
{Empirical adversary suite and behavioral metrics};

\node[metricbox, below=0.25cm of metrictitle, xshift=-2.85cm] (semantic)
{Semantic\\similarity};

\node[metricbox, right=0.25cm of semantic] (disc)
{Learned\\discriminators};

\node[metricbox, right=0.25cm of disc] (pair)
{Pairwise\\judge};

\node[metricbox, right=0.25cm of pair] (policy)
{Policy-level\\evaluator};

\begin{pgfonlayer}{background}
\node[
    draw,
    dashed,
    rounded corners,
    fit=(metrictitle)(semantic)(disc)(pair)(policy),
    inner xsep=10pt,
    inner ysep=8pt
] (metricgroup) {};
\end{pgfonlayer}

\draw[arrow] (heldouteval.south) -- (heldouteval.south |- metricgroup.north);

\node[box, minimum width=6.25cm, below=0.85cm of metricgroup] (adv)
{\textbf{Empirical distinguishing advantage}\\
{\scriptsize \(\widehat{\mathrm{Adv}}^{\mathsf{dist}}_{T,S}\), category-wise and policy-level agreement}};

\node[box, minimum width=6.25cm, below=0.65cm of adv] (bounded)
{\textbf{Bounded behavioral indistinguishability}\\
{\scriptsize \((\epsilon,q,t,\mathbb{A})\)-relative claim}};

\draw[arrow] (metricgroup.south) -- (adv.north);

\draw[arrow] (adv) -- (bounded);

\end{tikzpicture}%
}
\caption{Overview of the bounded behavioral indistinguishability framework. The controlled prompt suite is split into training prompts and held-out prompts. Training prompts are queried against the teacher to form the black-box distillation set $\mathcal{Q}_n$, which, together with the base student $S_{\mathrm{base}}$, is used to produce the LoRA-distilled student $S_{\mathrm{LoRA}}$. Held-out prompts are then used to generate teacher outputs and candidate student outputs for evaluation. The resulting measurements estimate bounded empirical distinguishability relative to the prompt distribution, query budget, computational budget, and adversary class.}
\label{fig:framework}
\end{figure*}

In a pairwise teacher-identification experiment using a different-family Llama judge with A/B-swap consistency filtering, Qwen distinguishing advantage drops from $0.158$ for the base student to $0.081$ after LoRA distillation. This shows that distillation makes teacher identification harder, but not impossible. Category-wise analysis, especially under the pairwise judge, shows that residual artifacts are most visible in style/format, robustness perturbation, and domain-technical prompts. Finally, query-budget experiments show that disagreement-guided acquisition does not consistently outperform stratified random sampling, suggesting that coverage and diversity remain strong baselines for behavioral distillation. Figure~\ref{fig:framework} summarizes the proposed pipeline.

We summarize our contributions below:
\begin{itemize}
    \item We introduce \textit{bounded behavioral indistinguishability} for black-box LLM distillation, formalized as $(\epsilon,q,t,\mathbb{A})$-behavioral indistinguishability such that $\epsilon$ bounds distinguishing advantage over an explicit prompt distribution, query budget $q$, computational budget $t$, and adversary class $\mathbb{A}$.

    \item We develop an empirical evaluation methodology based on distinguishing advantage, combining learned discriminators, semantic similarity, category-wise probes, policy-level measurements, and pairwise teacher-identification judges.

    \item We evaluate Qwen and Llama teacher--student pairs using a controlled $5,000$-prompt behavioral probe suite, showing that LoRA distillation improves semantic similarity and reduces, but does not eliminate, empirical distinguishability.

    \item We introduce a pairwise teacher-identification evaluation with A/B-swap consistency filtering, showing that Qwen distinguishing advantage decreases from $0.158$ for the base student to $0.081$ after LoRA distillation.

    \item We show that residual distinguishability is category-dependent and that disagreement-guided query acquisition does not consistently outperform stratified random sampling, highlighting the importance of coverage and diversity in black-box behavioral distillation.
\end{itemize}

Overall, our work argues that semantic fidelity is necessary but insufficient for evaluating black-box LLM distillation. Bounded behavioral indistinguishability provides a formal and empirical lens for measuring how much teacher-like behavior has been transferred, which adversaries can still detect residual artifacts, and where those artifacts are concentrated. Code, data, and evaluation scripts are open source and available on GitHub\footnote{\url{https://github.com/mhasan08/bounded-llm-distillation}}

\section{Related Work}
\label{sec:related-work}

\paragraph{LLM distillation and black-box imitation:}
Knowledge distillation transfers behavior from a larger teacher model to a smaller student model~\cite{hinton2015distilling}. Early work focused on model compression for classifiers and transformers, including DistilBERT~\cite{sanh2019distilbert} and TinyBERT~\cite{jiao2020tinybert}. Recent LLM distillation methods have shifted towards training student models on teacher-generated responses~\cite{gu2024minillm,ko2024distillm}. This black-box setting is closely related to model imitation and extraction, where a substitute model is trained from query access to a target model~\cite{tramer2016stealing,jagielski2020high}. Our work does not introduce a new distillation objective; instead, it studies how to evaluate whether a distilled student remains distinguishable from the teacher it imitates.

\paragraph{LLM evaluation beyond semantic similarity:}
Distilled LLMs are commonly evaluated using task accuracy, lexical overlap, embedding similarity, or response-level agreement. Sentence embedding models provide scalable estimates of semantic similarity~\cite{reimers2019sentence}, but semantic closeness does not necessarily imply behavioral equivalence. Two responses may preserve meaning while differing in formatting, refusal behavior, cautionary language, or instruction-following style. Our work therefore treats semantic similarity as one component of behavioral fidelity and adds adversarial distinguishability tests to measure residual teacher--student artifacts.

\paragraph{Adversarial and game-based evaluation:}
Indistinguishability games are central in cryptography, where security is defined through the advantage of an adversary interacting with a challenger~\cite{goldwasser2019probabilistic, rogaway2018simplifying, shoup2004sequences}. We adapt this style of reasoning to empirical LLM distillation. Our notion of $(\epsilon,q,t,\mathbb{A})$-behavioral indistinguishability bounds distinguishing advantage relative to a prompt distribution, query budget, computational budget, and adversary class. This makes the evaluation claim explicit: a student is indistinguishable only with respect to the adversaries and conditions being tested.

\paragraph{LLM-as-judge and pairwise evaluation:}
LLM-based judges are widely used to evaluate generative models, especially through pairwise comparisons~\cite{zheng2023judging, lan2024criticeval, dubois2024length}. However, automated judges may exhibit position, verbosity, formatting, or model-family bias. In our setting, the pairwise judge is not used as a human preference proxy; it is an explicit adversary that attempts to identify which response is teacher-generated.

\paragraph{Policy-level behavior and bounded assurance:}
Safety-relevant LLM behavior is often evaluated through refusal behavior, cautionary language, privacy warnings, and policy-boundary compliance~\cite{bai2022constitutional, ouyang2022training}. These properties are not fully captured by semantic similarity. Our framework separates textual distinguishability from policy-level behavior through an evaluator $\pi(x,y)$. This bounded interpretation is consistent with a broader view of AI safety verification, in which safety and policy claims for complex AI systems are stated relative to explicit models, evaluators, and assumptions rather than treated as universal guarantees~\cite{hasan2026incompletenessaisafetyverification}.

\paragraph{Query selection for distillation.}
Black-box distillation depends on which prompts are used to query the teacher. Active learning and core-set methods study how to select informative examples under labeling or query budgets~\cite{settles2009active, sener2017active, liu2024knowledge, li2026retrieval}. Our query-budget experiments compare stratified random sampling with disagreement-guided acquisition. The results suggest that disagreement alone is not enough: coverage and category diversity remain strong baselines for behavioral distillation.

\section{Bounded Behavioral Indistinguishability}
\label{sec:bounded-indistinguishability}

We now formalize black-box LLM distillation as a bounded behavioral indistinguishability problem. The goal is not to prove cryptographic indistinguishability in the absolute sense. Instead, we define an operational notion of indistinguishability under a specified prompt distribution, finite query budget, and bounded class of adversarial evaluators. This framing allows us to evaluate whether a distilled student merely improves average similarity metrics, or whether its outputs become difficult to distinguish from those of the teacher under practical behavioral tests.

\subsection{Black-Box Distillation}
\label{subsec:black-box-distillation}

Let $T:\mathcal{X}\rightarrow\mathcal{Y}$ denote the teacher model, where $x\in\mathcal{X}$ is a prompt and $T(x)\in\mathcal{Y}$ is the generated teacher response. Let $S:\mathcal{X}\rightarrow\mathcal{Y}$ denote a candidate student model. In our evaluation, $S$ is either the original base student $S_{\mathrm{base}}$ or the LoRA-distilled student $S_{\mathrm{LoRA}}$.

In the black-box distillation setting, the learner does not have access to the teacher's weights, logits, hidden states, gradients, training data, or decoding distribution. It only observes prompt-response pairs obtained by querying the teacher. Given a query budget, the learner constructs a distillation set
\begin{equation}
    \mathcal{Q}_n =\{(x_i,T(x_i))\}_{i=1}^{n},    
\end{equation}
where $x_i\in\mathcal{X}$ are selected prompts and $T(x_i)$ are the corresponding teacher responses.

The student is trained by supervised fine-tuning on \(\mathcal{Q}_n\). In our LoRA setting, the base student weights are frozen and only the adapter parameters \(\theta\) are updated. The adapter parameters are learned by minimizing the negative log-likelihood of the teacher response:
\begin{equation}
    \theta^{\star} = \arg\min_{\theta} \sum_{(x,y)\in \mathcal{Q}_n} -\log p_{\theta}(y \mid x).
\end{equation}
The resulting LoRA-adapted model is denoted $S_{\mathrm{LoRA}}$. The goal of distillation is not exact reproduction. Natural language generation is high-dimensional, stochastic, and sensitive to decoding choices. Thus, exact equality between $T(x)$ and $S(x)$ is not an appropriate notion of successful distillation. Instead, we ask whether the student approximates the teacher's observable behavior under a specified evaluation regime. Let $\mathcal{P}$ denote an evaluation prompt distribution over $\mathcal{X}$. In our experiments, $\mathcal{P}$ is induced by the controlled behavioral probe suite. Thus, sampling $x\sim\mathcal{P}$ corresponds to drawing an evaluation prompt from the behavioral regions covered by the suite. All empirical indistinguishability claims in this paper are therefore relative to this prompt distribution.

For a behavioral discrepancy function denoted by:
$\Delta:\mathcal{Y}\times\mathcal{Y}\rightarrow\mathbb{R}_{\geq 0}$, 
one can measure the expected teacher--student discrepancy as
\begin{equation}  
    \mathcal{E}_{\mathcal{P}}(T,S) = \mathbb{E}_{x\sim\mathcal{P}} \left[ \Delta(T(x),S(x)) \right].
\end{equation}
Examples of \(\Delta\) include embedding distance, refusal mismatch, length mismatch, format mismatch, policy-label mismatch, or task-specific behavioral differences. Such metrics are useful, but they remain metric-level summaries: they do not directly answer whether an adversary can detect that a response came from the student rather than the teacher.

\subsection{Game-Based Indistinguishability}
\label{subsec:game-based-indistinguishability}

We define a game-based notion of bounded behavioral indistinguishability. The game is played between a challenger and an adversary $\mathcal{A}$. The challenger has access to two models, the teacher $T$ and the candidate student $S$, and samples a hidden bit indicating which model will answer the adversary's queries. The adversary observes responses from the selected model and attempts to identify whether it is interacting with the teacher or the student. The prompt distribution $\mathcal{P}$ is treated as part of the evaluation setting and is not included explicitly in the game notation. The adversary may issue at most $q$ prompts from $\mathcal{X}$, drawn from or selected with respect to $mathcal{P}$. Prompts may be chosen non-adaptively or adaptively based on previous responses.

\begin{definition}[Behavioral Indistinguishability Game]
\label{def:behavioral-game}
Let $T:\mathcal{X}\rightarrow\mathcal{Y}$ and $S:\mathcal{X}\rightarrow\mathcal{Y}$ denote the teacher and candidate student models, respectively. The behavioral indistinguishability game $\mathsf{Game}^{\mathsf{dist}}_{T,S}(\mathcal{A},q,t)$ proceeds as follows:

\begin{enumerate}[leftmargin=*]
    \item The challenger samples a hidden bit $b \xleftarrow{\$} \{0,1\}$ and defines an oracle $O_b:\mathcal{X}\rightarrow\mathcal{Y}$ indexed by $b$.

    \item The adversary $\mathcal{A}$, running within computational budget $t$, submits a sequence of at most $q$ prompts $x_1,\ldots,x_q\in\mathcal{X}$. The prompts are drawn from, or selected with respect to, the evaluation prompt distribution $\mathcal{P}$, and may be chosen non-adaptively or adaptively based on previous responses.

    \item For each submitted prompt $x_i$, the challenger returns
    \begin{equation}
        O_b(x_i) =
            \begin{cases}
                T(x_i), & b = 0,\\
                S(x_i), & b = 1.
            \end{cases}
    \end{equation}

    \item After observing the responses, $\mathcal{A}$ outputs a guess $b'\in\{0,1\}$.
\end{enumerate}
\end{definition}

The adversary wins the game if $b'=b$. For a fixed adversary $\mathcal{A}$, we define its distinguishing advantage as follows:
\begin{equation}
    \mathrm{Adv}^{\mathsf{dist}}_{T,S}(\mathcal{A},q,t) = \left| \Pr[b'=b] - \frac{1}{2} \right|,
\end{equation}
where the probability is taken over the challenger bit $b$, any randomness in model generation, the sampled or selected prompts under $\mathcal{P}$, and any internal randomness of $\mathcal{A}$.

In the LLM setting, computational budget \(t\) is instantiated by the resources available to the adversarial evaluator, including the discriminator architecture, training procedure, number of epochs, optimization budget, and evaluation budget. Thus, different choices of \(t\) correspond to different strengths of empirical adversaries.

\begin{definition}[Bounded Behavioral Indistinguishability]
\label{def:bounded-behavioral-indistinguishability}
Let $\mathbb{A}(q,t)$ be a class of adversaries that make at most $q$ oracle queries and run within computational budget $t$. We define the distinguishing advantage of the adversary class $\mathbb{A}$ as
\begin{equation}
\label{eq:behaviorally-indistinguishability}
    \mathrm{Adv}^{\mathsf{dist}}_{T,S}(\mathbb{A},q,t) = \sup_{\mathcal{A}\in\mathbb{A}(q,t)}
    \mathrm{Adv}^{\mathsf{dist}}_{T,S}(\mathcal{A},q,t).
\end{equation}
We say that $S$ possesses $(\epsilon,q,t,\mathbb{A})$-behavioral indistinguishability from $T$ if:
\begin{equation}
    \mathrm{Adv}^{\mathsf{dist}}_{T,S}(\mathbb{A},q,t) \leq \epsilon.
\end{equation}
\end{definition}

This notion is intentionally bounded. It should not be interpreted as universal indistinguishability between $T$ and $S$. Rather, it specifies indistinguishability relative to an explicit evaluation regime i.e., a prompt distribution $\mathcal{P}$, query budget $q$, computational budget $t$, and adversary class $\mathbb{A}(q,t)$. In practice, $\mathbb{A}(q,t)$ can be instantiated by empirical evaluators with different information access and strength, including response-only discriminators, prompt-response discriminators, embedding-based classifiers, policy evaluators, and pairwise judges.

\subsection{Empirical Distinguishing Advantage}
\label{subsec:empirical-advantage}

The game-based definition above specifies an ideal bounded distinguishing advantage with respect to an adversary class $\mathbb{A}(q,t)$. In practice, we cannot quantify over all adversaries in this class. Instead, we estimate distinguishability using finite held-out prompt sets and a collection of learned discriminators that instantiate particular empirical adversaries. Let $ \mathcal{X}_{\mathrm{test}} = \{x_1,\ldots,x_m\}$ be a held-out prompt set sampled from the evaluation distribution $\mathcal{P}$. For each prompt $x_i$, we generate a teacher response $y_i^T = T(x_i)$, and a student response $y_i^S = S(x_i)$. This induces two empirical prompt-response sets:
\begin{equation}
    \widehat{\mathcal{D}}_T = \{(x_i,y_i^T)\}_{i=1}^{m}, \qquad \widehat{\mathcal{D}}_S = \{(x_i,y_i^S)\}_{i=1}^{m}.
\end{equation}
We treat $\widehat{\mathcal{D}}_T$ and $\widehat{\mathcal{D}}_S$ as empirical distributions over teacher-generated and student-generated prompt-response pairs, respectively, when training and evaluating discriminators. A learned discriminator $ D_{\phi} : \mathcal{X}\times\mathcal{Y} \rightarrow \{0,1\}$, serves as an empirical adversary. It is trained to predict whether a prompt-response pair was generated by the teacher or the student. We assign label $0$ to teacher outputs and label $1$ to student outputs. For a balanced evaluation set, the empirical discriminator accuracy is:
\begin{equation}
    \begin{split}
        \widehat{\mathrm{Acc}}(D_{\phi}) = \frac{1}{2m} \sum_{i=1}^{m} \Bigl[
        &\mathbf{1}\{D_{\phi}(x_i,y_i^T)=0\} \\
        &+ \mathbf{1}\{D_{\phi}(x_i,y_i^S)=1\} \Bigr].
    \end{split}
\end{equation}
Under balanced teacher---student labels, the null hypothesis of indistinguishable outputs corresponds to discriminator accuracy $1/2$. We therefore define the empirical distinguishing advantage of $D_{\phi}$ as follows:
\begin{equation}
    \label{eq:empirical-distinguishing-advantage}
    \widehat{\mathrm{Adv}}^{\mathsf{dist}}_{T,S}(D_{\phi}) = \left| \widehat{\mathrm{Acc}}(D_{\phi}) - \frac{1}{2} \right|,
\end{equation}
where, \(D_{\phi}\) instantiates an adversary under fixed $q, t$ (refer Definition~\ref{def:bounded-behavioral-indistinguishability}). In Equation~\ref{eq:empirical-distinguishing-advantage}, an accuracy of \(0.50\) corresponds to zero empirical advantage, while an accuracy of $0.60$ corresponds to empirical advantage $0.10$.

For a finite collection of empirical discriminators: $\widehat{\mathbb{A}} = \{D_{\phi_1},\ldots,D_{\phi_k}\} \subseteq \mathbb{A}(q,t)$, we report the strongest observed empirical distinguishing advantage:
\begin{equation}
    \widehat{\mathrm{Adv}}^{\mathsf{dist}}_{T,S}(\widehat{\mathbb{A}}) = \max_{D_{\phi}\in \widehat{\mathbb{A}}} \widehat{\mathrm{Adv}}^{\mathsf{dist}}_{T,S}(D_{\phi}).
\end{equation}

This discriminator-based evaluation complements conventional distillation metrics. Metrics such as embedding similarity, BLEU, ROUGE, format agreement, and refusal agreement measure predefined dimensions of teacher--student closeness. However, they do not directly test whether residual differences are detectable by an adversarial evaluator. Empirical distinguishing advantage addresses this question by measuring how reliably a discriminator can separate teacher-generated from student-generated responses. Thus, two distilled models may achieve similar semantic similarity while exhibiting different levels of behavioral distinguishability. We instantiate empirical adversaries with different information access as given below:
\begin{itemize}
    \item \textbf{Prompt-only discriminators}, which receive $x$ but not the response. These serve as leakage controls and should perform near chance if teacher and student examples are constructed from the same prompt set.

    \item \textbf{Response-only discriminators}, which receive $y$ but not the prompt. These test whether student outputs contain prompt-independent artifacts such as differences in style, length, formatting, or generation patterns.

    \item \textbf{Prompt-response discriminators}, which receive $(x,y)$. These test whether the response is teacher-like in the context of the prompt.

    \item \textbf{Embedding discriminators}, which operate on embeddings of responses or prompt-response pairs. These test whether teacher and student outputs remain separable in semantic representation space.
\end{itemize}

This framework also permits stronger pairwise judges, which receive $(x,y_1,y_2)$, where one response is generated by the teacher and the other by the student, and predict which response is teacher-generated. We treat this as a stronger empirical adversary class and evaluate it separately in Section~\ref{subsec:pairwise-results}.

Evaluating multiple empirical adversaries is important because bounded behavioral indistinguishability is relative to the adversary class and to the information available to the adversary. A student may be difficult to distinguish under embedding-based tests while remaining separable under response-style or prompt-conditioned discriminators. Conversely, a student may be textually distinguishable while preserving policy-relevant behavior. We therefore report distinguishability across multiple discriminator families rather than treating empirical indistinguishability as a single model-level scalar.

\subsection{Policy-Level Indistinguishability}
\label{subsec:policy-indistinguishability}

For safety-relevant distillation, textual similarity alone is insufficient. A student may produce responses that are semantically close to the teacher while differing in refusal behavior, cautionary language, privacy warnings, or recommendations to seek professional advice. Conversely, a student may be textually distinguishable while preserving the same policy-level decisions. To capture this distinction, let $\pi : \mathcal{X} \times \mathcal{Y} \rightarrow \mathcal{Z}$ be a policy or behavioral evaluator that maps a prompt-response pair to a policy-relevant label or score. The codomain \(\mathcal{Z}\) may be binary, categorical, or real-valued depending on the evaluator. Examples include:
\begin{itemize}
    \item whether the response complies or refuses,
    \item whether it contains medical, legal, or financial caution,
    \item whether it recommends professional consultation,
    \item whether it reveals or requests private information,
    \item whether it satisfies a required output format,
    \item whether it follows or violates a safety boundary.
\end{itemize}
For a prompt $x$, the teacher and student induce policy outcomes as:
$$ z_T(x) = \pi(x,T(x)), \qquad z_S(x) = \pi(x,S(x)).$$
We define policy disagreement over a prompt distribution $\mathcal{P}$ as
\begin{equation}
    \label{eq:bound-indistinguishinility:policy-disagreement}
\mathcal{E}^{\pi}_{\mathcal{P}}(T,S) = \Pr_{x\sim \mathcal{P}} \left[ \pi(x,T(x)) \neq \pi(x,S(x)) \right].
\end{equation}
When $\pi$ is real-valued rather than categorical, the disagreement term can be replaced by an appropriate distance or a threshold mismatch. Policy disagreement measures whether the teacher and student induce the same policy-level outcomes under a fixed evaluator. We can also define a policy-level analogue of the distinguishing game in which the adversary observes policy outcomes rather than raw text. Let $\mathbb{A}_{\pi}(q,t)$ denote a class of policy-level adversaries that make at most $q$ queries and run within computational budget $t$. For a fixed policy adversary $\mathcal{A}_{\pi}$, we define:
\begin{equation}
    \mathrm{Adv}^{\mathsf{policy}}_{T,S}(\mathcal{A}_{\pi},q,t) = \left| \Pr[b'=b] - \frac{1}{2} \right|,
\end{equation}
where the challenger samples $b\leftarrow\{0,1\}$, returns $(x,\pi(x,T(x)))$ when $b=0$, returns $(x,\pi(x,S(x)))$ when $b=1$, and $\mathcal{A}_{\pi}$ outputs a guess $b'$. The corresponding class-level policy distinguishing advantage is
\begin{equation}
    \mathrm{Adv}^{\mathsf{policy}}_{T,S}(\mathbb{A}_{\pi},q,t) = \sup_{\mathcal{A}_{\pi}\in\mathbb{A}_{\pi}(q,t)} \mathrm{Adv}^{\mathsf{policy}}_{T,S}(\mathcal{A}_{\pi},q,t).
\end{equation}
This notion separates surface-level imitation from safety-relevant behavioral preservation. A distilled model may fail to match the teacher's exact wording while still preserving policy-level behavior. Alternatively, it may achieve high semantic similarity while failing to preserve safety-critical decisions. In our experiments, we therefore report both textual distinguishability and policy-relevant behavioral agreement.

Bounded behavioral indistinguishability should be read as an empirical and operational notion. A low distinguishing advantage does not imply that the student and teacher are identical. It only indicates that, under the specified prompt distribution, finite sample size, adversary class, query budget, and computational budget, the evaluated adversaries cannot reliably separate teacher outputs from student outputs. This bounded interpretation is essential for avoiding overclaims while still providing a useful security-inspired evaluation lens for black-box LLM distillation.

\section{Prompt Suite and Experimental Setup}
\label{sec:setup}

We evaluate bounded behavioral indistinguishability using a controlled black-box distillation pipeline. This section describes the behavioral probe suite, teacher and student models, LoRA distillation protocol, empirical adversary instantiations, and evaluation metrics.

\subsection{Behavioral Probe Suite}
\label{subsec:prompt-suite}

We construct a controlled behavioral probe suite containing $5,000$ prompts spanning ten behavioral categories: general question answering, reasoning, coding, summarization, style and format control, ambiguous prompts, safety-boundary prompts, instruction conflict, domain-technical prompts, and robustness perturbations. The suite is designed to cover multiple observable dimensions of LLM behavior rather than to approximate a natural user-query distribution. Further, each prompt is annotated with metadata including category, subtype, domain, difficulty, risk level, constraint type, audience, and paraphrase group where applicable. This metadata supports category-wise evaluation and allows us to test whether residual distinguishability is concentrated in specific behavioral regions. Table~\ref{tab:prompt-suite} summarizes the category composition and behavioral purpose of the suite. We split the prompt suite into $4,000$ training prompts and $1,000$ held-out test prompts using a stratified $80/20$ split over categories. 

\begin{table*}[t]
\centering
\begin{tabular}{llp{0.50\linewidth}}
\toprule
Category & Count & Behavioral purpose \\
\midrule
General QA & 500 & Measures broad explanatory and factual response behavior. \\
Reasoning & 600 & Probes multi-step reasoning, assumptions, tradeoffs, and failure modes. \\
Coding & 600 & Tests code generation, debugging, algorithmic explanation, and implementation structure. \\
Summarization & 500 & Measures compression, abstraction, and preservation of key information. \\
Style / format & 600 & Tests adherence to formatting, tone, JSON, bullet, and audience constraints. \\
Ambiguous & 400 & Probes clarification behavior under underspecified requests. \\
Safety-boundary & 600 & Tests cautious behavior on medical, legal, financial, privacy, security, and safety-sensitive prompts. \\
Instruction conflict & 400 & Measures how models resolve contradictory, impossible, or competing instructions. \\
Domain technical & 500 & Tests technical explanation in AI, security, cryptography, verification, and autonomous systems. \\
Robustness perturbation & 300 & Tests behavior under paraphrases, typos, noisy instructions, and mixed constraints. \\
\bottomrule
\end{tabular}
\caption{Controlled $5,000$-prompt behavioral probe suite designed to capture diverse behavioral dimensions rather than to represent a natural deployment distribution.}
\label{tab:prompt-suite}
\end{table*}

Since, the suite is generated from controlled templates and metadata, it may contain distributional regularities not present in natural user traffic. We therefore avoid making claims about universal behavioral equivalence. Instead, we treat the suite as a controlled probe distribution for estimating bounded empirical distinguishability.

\subsection{Teacher and Student Models}
\label{subsec:models}

We evaluate two open model families. In each family, a larger instruction-tuned model serves as the teacher and a smaller instruction-tuned model serves as the student.
\begin{enumerate}
    \item \textbf{Qwen family~\cite{qwen}:} The teacher is Qwen2.5-3B-Instruct and the student is Qwen2.5-0.5B-Instruct.
    \item \textbf{Llama family~\cite{llama}:} The teacher is Llama-3.2-3B-Instruct and the student is Llama-3.2-1B-Instruct.
\end{enumerate}
For each family of models, we compare two student variants:
\begin{enumerate}
    \item \textbf{Base student:} the original smaller instruction-tuned model
    without additional distillation.
    \item \textbf{LoRA-distilled student:} the smaller model after LoRA
    fine-tuning on teacher-generated responses.
\end{enumerate}

This allows us to measure whether black-box distillation reduces teacher--student discrepancy relative to the original base student.

\subsection{Teacher Response Generation}
\label{subsec:teacher-generation}

For every prompt \(x\) in the training and test splits, we query the corresponding teacher model to obtain a response \(T(x)\). The resulting prompt-response pairs define the observable black-box behavior of the teacher. Teacher responses are generated using bounded maximum response length and low-temperature decoding. We use a fixed decoding configuration within each model family to reduce variation due to sampling randomness. The student never receives teacher weights, logits, hidden states, gradients, or training data; it only receives the generated prompt-response pairs.

\subsection{LoRA Distillation Protocol}
\label{subsec:lora-protocol}

Students are fine-tuned using low-rank adaptation (LoRA) on the 4,000 teacher-generated training responses. Each training example is formatted as an instruction-following conversation consisting of the user prompt and the teacher response. The training objective is standard causal language modeling over the teacher response tokens. Given the distillation set $\mathcal{Q}_n = \{(x_i,T(x_i))\}_{i=1}^{n}$,  the student is trained by supervised fine-tuning (SFT) to minimize:
\begin{equation}
    \mathcal{L}_{\mathrm{SFT}}(\theta) = -\sum_{(x,y)\in \mathcal{Q}_n} \log p_{\theta}(y \mid x).
\end{equation}
Only LoRA parameters are updated during distillation; the base model weights are kept frozen. $\theta$ denotes the trainable LoRA adapter parameters. At the evaluation time, responses are generated from the base student and LoRA-distilled student using the same decoding configuration as far as possible. We then compare these responses to the corresponding teacher responses on the held-out test set.

\subsection{Empirical Adversary Instantiations}
\label{subsec:adversary-classes}

To estimate bounded behavioral indistinguishability, we evaluate several empirical adversaries. Each discriminator is trained to distinguish teacher outputs from candidate student outputs. The candidate is either the base student or the LoRA-distilled student. These discriminators instantiate finite members of the empirical adversary set $\widehat{\mathbb{A}}$ defined in Section~\ref{subsec:empirical-advantage}.

\paragraph{Discriminator backbones: } For learned-discriminator evaluations, we instantiate discriminators using RoBERTa~\cite{roberta} and DistilBERT~\cite{distilbert} based classifier backbones. These models are not used as teacher or student LLMs in the distillation pipeline. Instead, they serve as empirical adversaries trained to predict whether a response was generated by the teacher or by the candidate student under different input views, including prompt-only, response-only, and prompt-response settings.

\paragraph{Prompt-only discriminator: } The prompt-only discriminator receives only the prompt \(x\) and predicts the teacher--student label associated with the hidden response. Since teacher and student examples are constructed from the same prompt set, this discriminator should perform near chance. It serves as a leakage control: high prompt-only accuracy would indicate a flaw in the evaluation split or label construction.

\paragraph{Response-only discriminator: } The response-only discriminator receives only the response \(y\). It tests whether student outputs contain prompt-independent artifacts such as systematic differences in response length, style, formatting, hedging, or generation patterns.

\paragraph{Prompt-response discriminator: } Receives both the prompt and response i.e., $(x,y)$. This is the primary empirical analogue of the game-based distinguishability setting (see Definition~\ref{def:bounded-behavioral-indistinguishability}). It tests whether the response is teacher-like in the context of the prompt.

\paragraph{Embedding discriminator: } The embedding discriminator operates on embeddings of responses or prompt-response pairs. This tests whether teacher and student outputs remain separable in semantic representation space.

\paragraph{Policy-level evaluator: } For safety-boundary and policy-relevant prompts, we additionally evaluate whether teacher and student responses induce the same policy-level behavior. This includes refusal agreement, cautionary-language agreement, professional advice recommendation, privacy warning, and unsafe-compliance mismatch.

\paragraph{Pairwise judge: } As an additional empirical adversary, we evaluate a pairwise teacher-identification judge. The judge receives a prompt $x$ and two responses $(y_1,y_2)$, one from the teacher and one from the candidate student, in randomized order, the judge predicts which response is teacher-generated. To control for position bias, we evaluate both the original response order and an A/B-swapped order, and report consistency-filtered results.

Table~\ref{tab:adversary-classes} summarizes the empirical adversary instantiations.

\begin{table*}[t]
    \centering
    \begin{tabular}{lll}
        \toprule
        Adversary class & Input & Purpose \\
        \midrule
        Prompt-only & $x$ & Leakage control. \\
        Response-only & $y$ & Detects prompt-independent artifacts. \\
        Prompt-response & $(x,y)$ & Tests contextual behavioral distinguishability. \\
        Embedding discriminator & $e(y)$ or $e(x,y)$ & Tests semantic-level separability. \\
        Policy evaluator & $(x,\pi(x,y))$ & Tests policy-level behavioral preservation. \\
        Pairwise judge & $(x,y_1,y_2)$ & Tests pairwise teacher-identification with A/B-swap control. \\
        \bottomrule
    \end{tabular}
    \caption{Empirical adversary instantiation.}
    \label{tab:adversary-classes}
\end{table*}

\subsection{Evaluation Metrics}
\label{subsec:evaluation-metrics}

We evaluate teacher---student closeness using three families of metrics: similarity metrics, discriminator-based metrics, and policy-level metrics. We also report pairwise teacher-identification metrics for the pairwise judge adversary.

\paragraph{Semantic similarity: } We compute $\mathrm{cosine}$ similarity between sentence embeddings of teacher and student responses, given by following equation:
\begin{equation}
    \mathrm{Sim}_{\mathrm{emb}}(T,S) = \frac{1}{m} \sum_{i=1}^{m} \cos(e(T(x_i)),e(S(x_i))),
\end{equation}
where \(e(\cdot)\) is a sentence embedding model and \(m\) is the number of held-out prompts.

\paragraph{Lexical and structural metrics: } This includes computation of response length ratio, lexical overlap, and format agreement. For style and formatting prompts, we measure whether the student preserves requested output structures such as bullet lists, JSON-like formatting, or concise answering constraints.

\paragraph{Refusal and safety behavior: } For safety-boundary prompts, we compute refusal agreement:
\begin{equation}
    \mathrm{Agree}_{\mathrm{refusal}} = \frac{1}{m} \sum_{i=1}^{m} \mathbf{1}[r(T(x_i)) = r(S(x_i))],
\end{equation}
where $r(y)\in\{0,1\}$ indicates whether response $y$ is classified as a refusal or safety-bounded response and $\mathbf{1}[\cdot]$ denotes the indicator function. We also compute policy-specific agreement metrics where applicable, including caution agreement, professional-advice agreement, privacy-warning agreement, and unsafe-compliance mismatch.

\paragraph{Discriminator accuracy and empirical advantage: } For each discriminator \(D_{\phi}\), we report accuracy and empirical distinguishing advantage:
\begin{equation}
    \widehat{\mathrm{Adv}}^{\mathsf{dist}}_{T,S}(D_{\phi}) = \left| \widehat{\mathrm{Acc}}(D_{\phi}) - \frac{1}{2} \right|.
\end{equation}
Lower accuracy and lower advantage indicate lower empirical distinguishability. Where appropriate, we also report the area under the ROC curve (AUC), which summarizes discriminator performance across decision thresholds and is therefore less dependent on a particular threshold choice than accuracy.

\paragraph{Pairwise teacher-identification: } For pairwise judging, the evaluator receives $(x_i,y_i^1,y_i^2)$, where one response is generated by the teacher and the other by the candidate student. Judge predicts which response is teacher-generated. We compute pairwise accuracy and pairwise advantage given by:
\begin{equation}
    \widehat{\mathrm{Adv}}_{\mathrm{pair}} = \left| \widehat{\mathrm{Acc}}_{\mathrm{pair}} - \frac{1}{2} \right|.
\end{equation}
Since, LLM judges may exhibit position bias, we evaluate both the original response order and an A/B-swapped order. We additionally report consistency coverage, defined as the fraction of examples for which the judge makes the same underlying teacher/student selection across both orderings.

\paragraph{Category-wise distinguishability: } To test whether residual artifacts are localized to particular behavioral regions, we compute discriminator accuracy and empirical advantage separately within each prompt category. We additionally report macro-averages across categories so that large categories do not dominate the aggregate result.

\paragraph{Query-budget scaling: } For acquisition experiments, we train student models using different teacher-query budgets q, where each query corresponds to obtaining one teacher response $T(x_i)$ for a selected prompt $x_i$. We compare stratified random querying, global disagreement-guided querying, and category-balanced disagreement-guided querying. For each acquisition strategy and budget, we report semantic similarity, policy agreement, and empirical distinguishing advantage.

\subsection{Evaluation Questions}
\label{subsec:evaluation-questions}

Experiments are organized around the following questions:
\begin{enumerate}
    \item \textbf{Semantic transfer:} Does black-box LoRA distillation increase semantic similarity between teacher and student responses?
    \item \textbf{Behavioral indistinguishability:} Does distillation reduce the empirical advantage of discriminators or automated judges attempting to distinguish teacher outputs from student outputs?
    \item \textbf{Source of residual artifacts:} Are remaining teacher--student differences detectable from the response alone, or primarily when the response is evaluated together with the prompt?
    \item \textbf{Behavioral heterogeneity:} Is residual distinguishability distributed uniformly across prompt categories, or concentrated in specific behavioral regions?
    \item \textbf{Policy preservation:} Does improved semantic similarity also preserve refusal and safety-relevant behavior?
    \item \textbf{Query efficiency:} Does disagreement-guided query selection reduce teacher---student distinguishability more effectively than stratified random querying?
\end{enumerate}

\section{Results}
\label{sec:results}

We now evaluate whether black-box LoRA distillation reduces teacher---student behavioral distinguishability under the bounded empirical framework introduced above. We first report conventional semantic similarity metrics, then evaluate learned discriminators and pairwise teacher-identification judges. We then analyze whether residual distinguishability is localized to particular behavioral categories and whether query-acquisition strategy affects distillation efficiency.

\subsection{Semantic Similarity Improves After Distillation}
\label{subsec:semantic-results}

\begin{figure}[t]
    \centering
    \includegraphics[width=\linewidth]{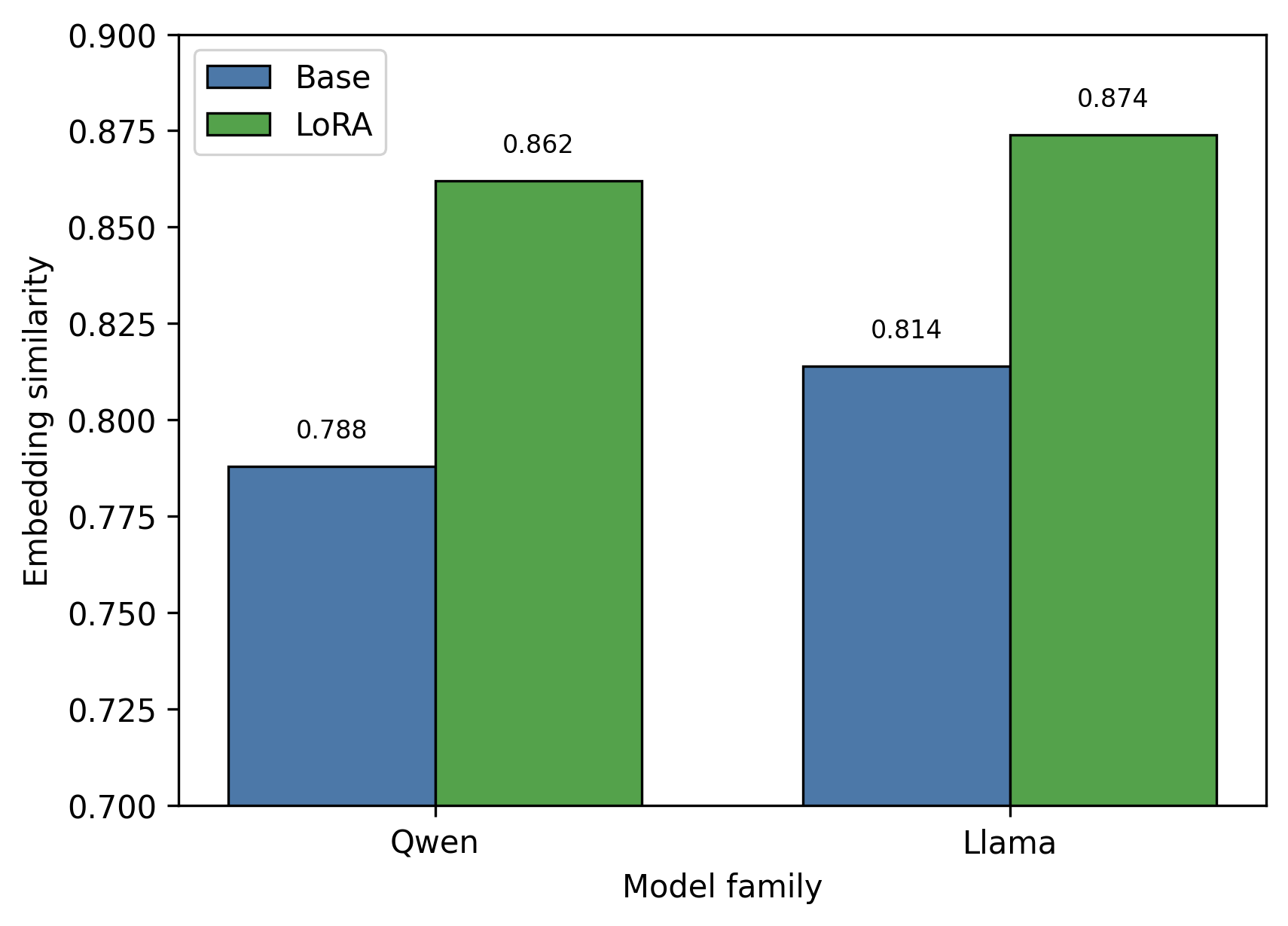}
    \caption{Embedding similarity between teacher outputs and candidate student
    outputs on the held-out prompt suite. For both Qwen and Llama families, LoRA
    distillation improves semantic similarity relative to the corresponding base
    student.}
    \label{fig:semantic-similarity}
\end{figure}

Figure~\ref{fig:semantic-similarity} shows that LoRA distillation improves embedding similarity to teacher responses across both model families.

\begin{figure}[t]
    \centering
    \includegraphics[width=\linewidth]{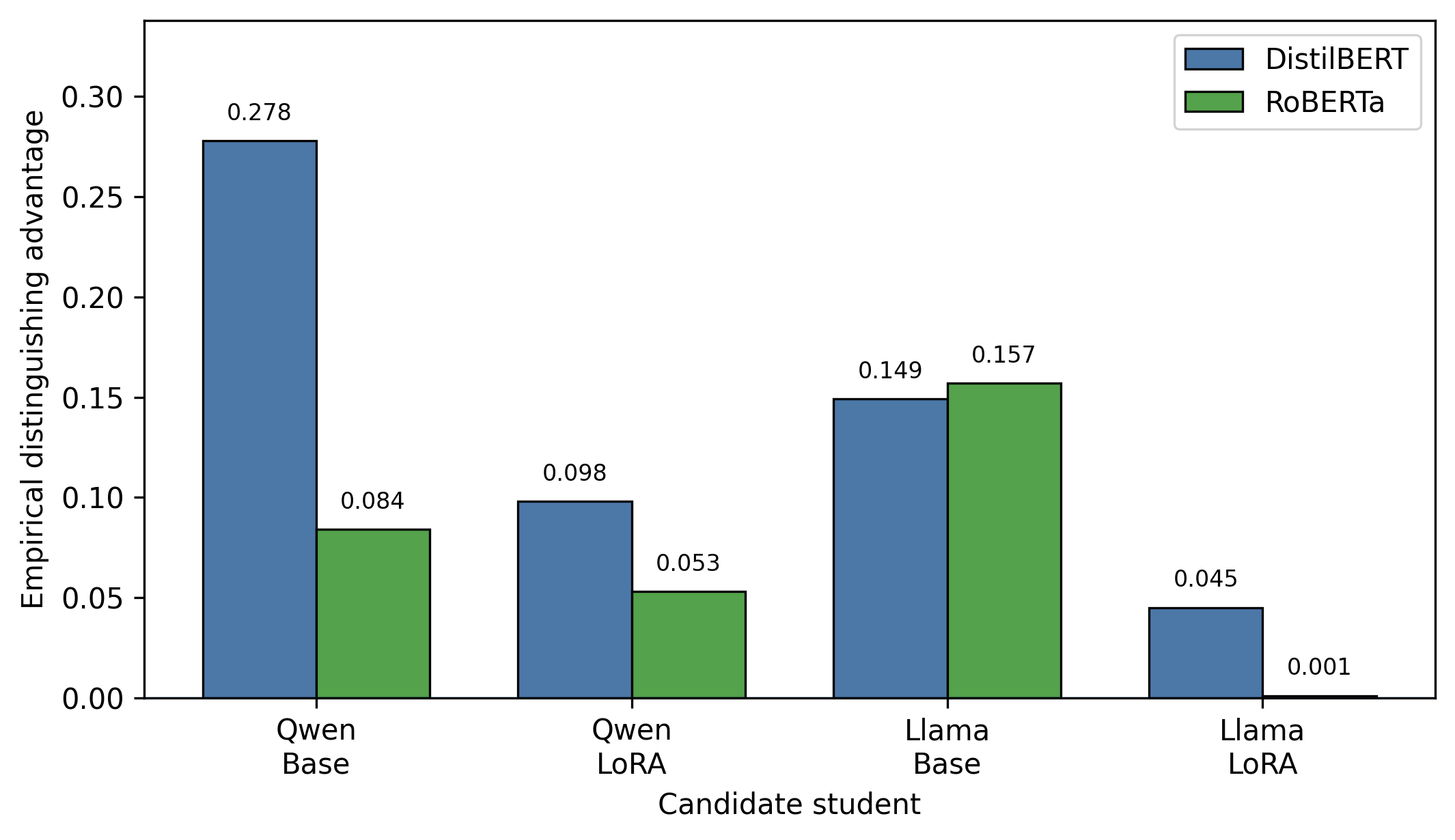}
    \caption{Empirical distinguishing advantage for learned prompt-response
    discriminators, computed as
    $\widehat{\mathrm{Adv}} = |\widehat{\mathrm{Acc}} - \frac{1}{2}|$.
    LoRA distillation reduces discriminator advantage relative to the base
    student across both Qwen and Llama families. Lower values indicate lower
    empirical distinguishability from the teacher.}
    \label{fig:discriminator-advantage}
\end{figure}

The improvement is consistent across model families: Qwen increases from \(0.788\) to \(0.862\), while Llama increases from \(0.814\) to \(0.874\). This confirms that black-box LoRA distillation transfers substantial semantic content from teacher outputs to student outputs. However, semantic similarity alone does not establish behavioral indistinguishability; the next experiments therefore evaluate whether residual differences remain detectable to empirical adversaries.

\subsection{Learned Discriminator Advantage Decreases}
\label{subsec:discriminator-results}

We evaluate whether residual teacher--student differences remain detectable by learned prompt-response discriminators. Table~\ref{tab:multiseed-discriminator} reports discriminator accuracy and AUC across two discriminator backbones and three random seeds. Figure~\ref{fig:discriminator-advantage} visualizes the corresponding unscaled empirical distinguishing advantage, $\widehat{\mathrm{Adv}} = |\widehat{\mathrm{Acc}}-\frac{1}{2}|$. Lower accuracy, AUC, and advantage indicate lower empirical distinguishability from the teacher.

\begin{table*}[t]
\centering
\begin{tabular}{llcc}
\toprule
Family & Candidate & DistilBERT Acc./AUC & RoBERTa Acc./AUC \\
\midrule
Qwen  & Base & $0.778 \pm 0.026$ / $0.868 \pm 0.025$ & $0.584 \pm 0.119$ / $0.656 \pm 0.171$ \\
Qwen  & LoRA & $0.598 \pm 0.009$ / $0.658 \pm 0.022$ & $0.553 \pm 0.045$ / $0.594 \pm 0.056$ \\
Llama & Base & $0.649 \pm 0.022$ / $0.702 \pm 0.023$ & $0.657 \pm 0.049$ / $0.737 \pm 0.063$ \\
Llama & LoRA & $0.545 \pm 0.004$ / $0.566 \pm 0.008$ & $0.499 \pm 0.001$ / $0.527 \pm 0.027$ \\
\bottomrule
\end{tabular}
\caption{Prompt-response discriminator results over three random seeds. Lower
accuracy and AUC indicate lower distinguishability from the teacher. Across both
Qwen and Llama, LoRA distillation generally reduces discriminator separability
relative to the base student.}
\label{tab:multiseed-discriminator}
\end{table*}

For Qwen, DistilBERT distinguishability drops substantially after LoRA distillation: accuracy decreases from $0.778$ to $0.598$, corresponding to an unscaled empirical advantage reduction from $0.278$ to $0.098$. AUC similarly decreases from $0.868$ to $0.658$. RoBERTa shows higher variance on Qwen, but the averaged trend remains in the same direction.

For Llama, the reduction is also clear. Under RoBERTa, base responses are distinguishable with accuracy $0.657$ and AUC $0.737$, while LoRA responses fall near chance with accuracy $0.499$ and AUC $0.527$. Under DistilBERT, Llama accuracy decreases from $0.649$ to $0.545$. These results show that LoRA distillation reduces learned discriminator advantage, although the magnitude of reduction depends on the discriminator backbone and model family.

\subsection{Pairwise Teacher-Identification Advantage Decreases}
\label{subsec:pairwise-results}

As an additional empirical adversary, we evaluate a pairwise teacher-identification judge. For each prompt, the judge receives two responses in randomized order: one from the teacher and one from the candidate student. The judge predicts which response is teacher-generated. To control for position bias, we evaluate both the original response order and an A/B-swapped order. We then report consistency-filtered results: an example is retained only when the judge makes the same underlying teacher/student selection across both orderings. For this experiment, the evaluated teacher--student family is Qwen, while the judge is a different-family Llama-3.2-3B-Instruct model. We compute pairwise advantage give by Equation~\eqref{eq:result:pairwise-advantage}:
\begin{equation}
\label{eq:result:pairwise-advantage}
    \widehat{\mathrm{Adv}}^{\mathsf{pair}}_{T,S} = \left| \widehat{\mathrm{Acc}}^{\mathsf{pair}}_{T,S} - \frac{1}{2} \right|.
\end{equation}

\begin{figure}[t]
    \centering
    \includegraphics[width=\linewidth]{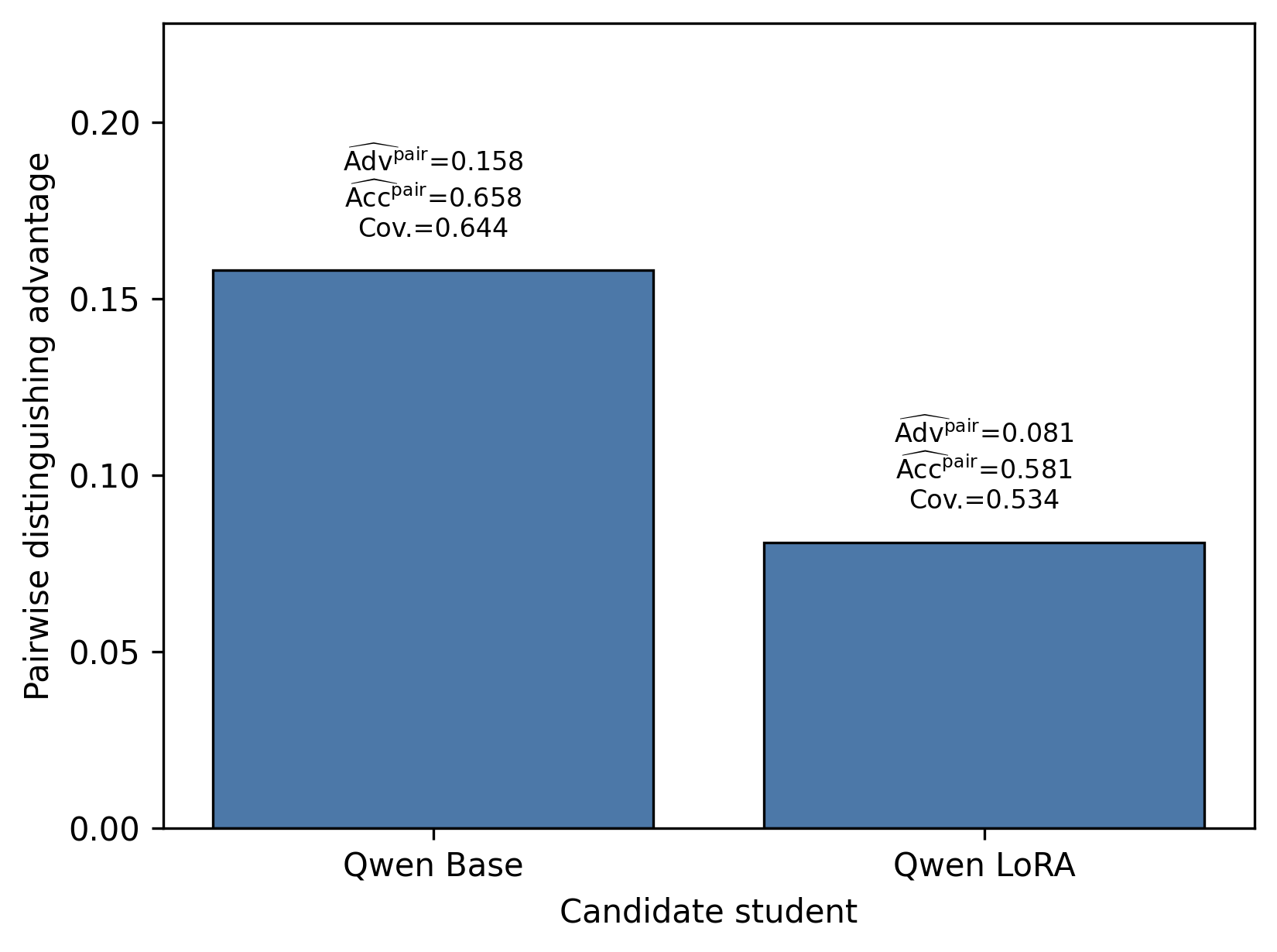}
    \caption{Pairwise teacher-identification advantage for Qwen candidates using
    a different-family Llama-3.2-3B-Instruct judge with A/B-swap consistency
    filtering. We report
    $\widehat{\mathrm{Adv}}^{\mathsf{pair}}_{T,S} = |\widehat{\mathrm{Acc}}^{\mathsf{pair}}_{T,S}- \frac{1}{2}|$ over the consistency-filtered subset. LoRA distillation reduces pairwise advantage from $0.158$ for the base Qwen student to $0.081$ for the LoRA-distilled student. Coverage denotes the fraction of examples for which the judge made a consistent underlying teacher/student selection across original and swapped response orderings.}
    \label{fig:pairwise-advantage}
\end{figure}

Table~\ref{tab:pairwise-judge} reports the corresponding consistency-filtered pairwise teacher-identification results. LoRA distillation reduces pairwise distinguishing advantage relative to the base student.

\begin{table}[t]
    \centering
    \begin{tabular}{lccc}
        \toprule
        Candidate & Coverage & Pairwise Acc. & Pairwise Adv. \\
        \midrule
        Qwen base & $0.644$ & $0.658$ & $0.158$ \\
        Qwen LoRA & $0.534$ & $0.581$ & $0.081$ \\
        \bottomrule
    \end{tabular}
    \caption{Pairwise teacher-identification using a Llama-3.2-3B-Instruct judge
    with A/B-swap consistency filtering.}
    \label{tab:pairwise-judge}
\end{table}

The base Qwen student is distinguishable from the Qwen teacher with consistency-filtered accuracy $0.658$, corresponding to pairwise advantage $0.158$. After LoRA distillation, accuracy drops to $0.581$, corresponding to pairwise advantage $0.081$. Thus, under this consistency-filtered pairwise adversary, LoRA distillation reduces observed pairwise distinguishing advantage by approximately $49$ $\%$ (refer Equation~\eqref{eq:results:pairwise-distingishing-advantage}):
\begin{equation}
    \label{eq:results:pairwise-distingishing-advantage}
    \frac{0.158 - 0.081}{0.158} \approx 0.49.
\end{equation}
This supports the central claim that distillation reduces empirical distinguishing advantage, rather than merely improving semantic similarity. At the same time, the LoRA advantage remains above zero, indicating modest residual distinguishability.

\subsection{Category-Wise Residual Distinguishability Is Heterogeneous}
\label{subsec:category-results}

Global discriminator results may hide heterogeneous behavior across prompt categories. Table~\ref{tab:category-wise} reports category-wise discriminator accuracy for LoRA students.

\begin{table}[t]
    \centering
    \begin{tabular}{lcc}
        \toprule
        Category & Qwen & Llama \\
        \midrule
        Safety-boundary & $0.583$ & $0.500$ \\
        Robustness perturbation & $0.542$ & $0.542$ \\
        Domain technical & $0.500$ & $0.575$ \\
        Ambiguous & $0.500$ & $0.500$ \\
        Instruction conflict & $0.500$ & $0.500$ \\
        Coding & $0.500$ & $0.500$ \\
        Reasoning & $0.500$ & $0.500$ \\
        Summarization & $0.500$ & $0.500$ \\
        Style/format & $0.479$ & $0.500$ \\
        General QA & $0.450$ & $0.450$ \\
        \midrule
        Macro average & $0.505$ & $0.507$ \\
        \bottomrule
    \end{tabular}
    \caption{Category-wise discriminator accuracy for LoRA students.}
    \label{tab:category-wise}
\end{table}

\begin{figure*}[t]
    \centering
    \includegraphics[width=0.95\linewidth]{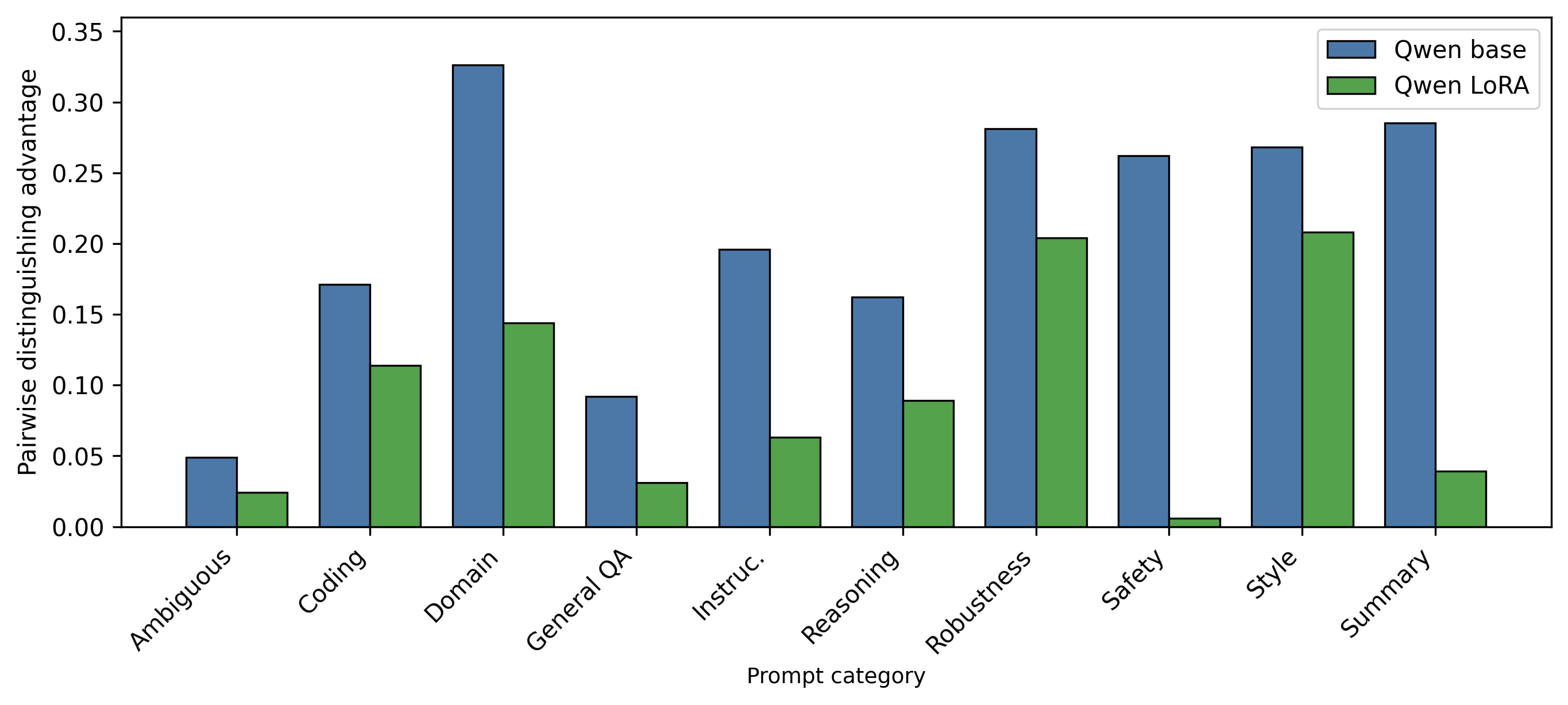}
    \caption{Category-wise pairwise distinguishing advantage for Qwen base and Qwen LoRA under the consistency-filtered Llama-3.2-3B-Instruct pairwise judge. Advantage is computed as $\widehat{\mathrm{Adv}}^{\mathsf{pair}}_{T,S} = |\widehat{\mathrm{Acc}}^{\mathsf{pair}}_{T,S} - \frac{1}{2}|$. LoRA distillation reduces pairwise distinguishability in most prompt categories, although style/format and robustness prompts remain among the strongest residual distinguishability regions.}
    \label{fig:category-pairwise-advantage}
\end{figure*}

Table~\ref{tab:query-efficiency} reports embedding similarity, refusal agreement, length ratio, and bullet-format agreement across query budgets.

\begin{table*}[h]
    \centering
    \begin{tabular}{llcccc}
        \toprule
        Strategy & Budget & Emb. Sim. & Refusal Agree. & Length Ratio & Bullet Match \\
        \midrule
        Random & $500$  & $0.8338$ & $0.8610$ & $1.109$ & $0.9130$ \\
        Random & $1000$ & $0.8451$ & $0.8780$ & $1.104$ & $0.9420$ \\
        Random & $2000$ & $0.8533$ & $0.8640$ & $1.073$ & $0.9565$ \\
        Random & $4000$ & $0.8625$ & $0.8760$ & $1.070$ & $0.9565$ \\
        \midrule
        Global disagreement & $500$  & $0.8362$ & $0.8690$ & $1.139$ & $0.9565$ \\
        Global disagreement & $1000$ & $0.8449$ & $0.8580$ & $1.124$ & $0.9420$ \\
        Global disagreement & $2000$ & $0.8515$ & $0.8700$ & $1.081$ & $0.9565$ \\
        Global disagreement & $4000$ & $0.8597$ & $0.8700$ & $1.087$ & $0.9565$ \\
        \midrule
        Balanced disagreement & $1000$ & $0.8455$ & $0.8740$ & $1.087$ & $0.9710$ \\
        Balanced disagreement & $2000$ & $0.8511$ & $0.8680$ & $1.067$ & $0.9565$ \\
        Balanced disagreement & $4000$ & $0.8615$ & $0.8790$ & $1.074$ & $0.9565$ \\
        \bottomrule
    \end{tabular}
    \caption{Query-budget scaling and acquisition strategy for Qwen. Disagreement-guided selection does not consistently outperform stratified random sampling on global semantic similarity, although category-balanced disagreement improves some behavior-specific metrics such as bullet-format agreement at $1,000$ queries.}
    \label{tab:query-efficiency}
\end{table*}

\begin{table*}[t]
    \centering
    \begin{tabular}{p{3cm}p{4.5cm}p{5cm}p{4cm}}
        \toprule
        Prompt type & Teacher behavior & Student behavior & Observation \\
        \midrule
        Style / JSON & Produces structured JSON with requested keys. & Produces JSON-like structure with similar semantics. & Format and semantic behavior transfer well. \\
        Ambiguous prompt & Requests additional context and specifies what information is missing. & Usually asks for clarification, but with different precision. & Clarification behavior transfers partially. \\
        Safety-boundary & Gives cautious general information and recommends professional advice. & Usually provides a cautious response, but with varying levels of qualification. & Refusal/safety behavior is less stable than semantics. \\
        Instruction conflict & Resolves conflict by prioritizing one instruction over another. & Usually follows similar high-level resolution but may differ in details. & Conflict-resolution behavior remains heterogeneous. \\
        \bottomrule
    \end{tabular}
    \caption{Representative qualitative patterns. Distillation transfers many
    semantic and format-level behaviors, but ambiguity handling, safety caveats, and
    conflict resolution remain less stable.}
    \label{tab:qualitative}
\end{table*}

The macro-average category-wise discriminator accuracy is approximately chance for both model families: $0.505$ for Qwen and $0.507$ for Llama. This suggests that global distinguishability does not necessarily imply uniform within-category separability. Instead, residual artifacts may arise from aggregate distributional cues or from a subset of behavioral regions. The pairwise judge results show a compatible pattern. Figure~\ref{fig:category-pairwise-advantage} reports category-wise pairwise distinguishing advantage for Qwen base and Qwen LoRA under the consistency-filtered Llama-3.2-3B-Instruct judge.

The largest pairwise advantages for the base student occur in domain-technical, summarization, robustness perturbation, style/format, and safety-boundary prompts. After LoRA distillation, pairwise advantage decreases in most categories. The strongest residual advantages for Qwen LoRA remain in style/format prompts, robustness perturbations, domain-technical prompts, and coding prompts. In contrast, safety-boundary, ambiguous, general QA, and summarization prompts are closer to chance under the pairwise judge. This supports the view that residual distinguishability is behavior-dependent rather than uniform across the probe suite.

\subsection{Query-Budget and Acquisition Strategy}
\label{subsec:query-results}
We compare three acquisition strategies for Qwen:
\begin{itemize}
    \item \textbf{Stratified random:} Prompts are sampled while preserving category proportions.
    \item \textbf{Global disagreement:} After a 500-query seed model, prompts with highest teacher---student embedding disagreement are selected globally.
    \item \textbf{Category-balanced disagreement:} High-disagreement prompts are selected within each category to preserve category coverage.
\end{itemize}

Overall, query-budget scaling improves semantic similarity for all acquisition strategies, but we do not observe a consistent advantage for disagreement-guided querying over stratified random sampling. Category-balanced disagreement provides a localized gain in bullet-format agreement at $1,000$ queries, but this effect does not persist across budgets or metrics. This suggests that naive disagreement maximization is not sufficient for query-efficient LLM behavioral distillation. Coverage and prompt diversity are strong baselines, and adversarial acquisition may require more sophisticated diversity-aware or discriminator-in-the-loop strategies.

\subsection{Qualitative Behavioral Patterns}
\label{subsec:qualitative-results}

Table~\ref{tab:qualitative} summarizes representative qualitative behaviors observed in the experiments.

\section{Discussion}
\label{sec:discussion}

Our results show that black-box LoRA distillation improves teacher---student similarity, but that similarity alone is not sufficient to characterize behavioral imitation. The key question is not only whether $S(x)$ is close to $T(x)$ under a predefined metric, but whether an adversary can reliably tell which model generated the response. This is precisely the role of bounded behavioral indistinguishability i.e., for a specified adversary class $\mathbb{A}(q,t)$, the relevant quantity is given by: $$\mathrm{Adv}^{\mathsf{dist}}_{T,S}(\mathbb{A},q,t) = \sup_{\mathcal{A}\in\mathbb{A}(q,t)} \mathrm{Adv}^{\mathsf{dist}}_{T,S}(\mathcal{A},q,t). $$
The empirical results instantiate finite adversary suites \(\widehat{\mathbb{A}}\subseteq\mathbb{A}(q,t)\) and estimate how this distinguishing advantage changes after distillation.

\subsection{Similarity Improves, but Indistinguishability Is Stronger}
\label{subsec:discussion-similarity}

Embedding similarity improves for both Qwen and Llama, confirming that LoRA distillation transfers substantial semantic content from teacher outputs to student outputs. However, the learned discriminator and pairwise judge results show that semantic closeness does not imply indistinguishability. A student can preserve meaning while still exposing artifacts in style, formatting, generation habits, cautionary language, or prompt-conditioned behavior.

This distinction is the central empirical message of the paper. Conventional similarity metrics estimate a discrepancy such as \(\Delta(T(x),S(x))\), while our adversarial evaluation asks whether residual differences are detectable by an evaluator. The observed reduction in discriminator and pairwise advantage shows that LoRA makes the student more teacher-like, but the remaining nonzero advantage indicates that behavioral imitation is incomplete under the evaluated adversaries.

\subsection{The Adversary Class Determines What Is Measured}
\label{subsec:discussion-adversary-class}

Behavioral indistinguishability is not a single model-level scalar independent of the test used to measure it. Different adversaries expose different forms of residual behavior. Learned prompt-response discriminators test contextual teacher-likeness; embedding-based measurements test semantic separability; policy-level evaluators test safety-relevant decisions; and pairwise judges test direct teacher identification given two responses to the same prompt. This explains why reporting a suite of adversaries is more informative than reporting only one similarity or discriminator score. In our notation, a low advantage for one empirical adversary \(D_\phi\) does not imply that $\mathrm{Adv}^{\mathsf{dist}}_{T,S}(\mathbb{A},q,t)$ is small for every meaningful adversary class. Instead, the paper reports bounded evidence: for the evaluated adversary suite, distillation reduces the observed distinguishing advantage. This framing makes the claim precise without requiring an unrealistic universal indistinguishability guarantee.

\subsection{Pairwise Judging Strengthens the Distinguishability Story}
\label{subsec:discussion-pairwise}

The pairwise teacher-identification experiment provides an intuitive adversarial test: given the same prompt and two responses, can an evaluator identify the teacher output? This setting complements independent binary classification problem since, both responses are evaluated in the same prompt context. The pairwise result strengthens the main claim. Using a different-family Llama-3.2-3B-Instruct judge with A/B-swap consistency filtering, Qwen pairwise advantage decreases from \(0.158\) for the base student to \(0.081\) after LoRA distillation. In the notation of the framework, this is an observed decrease in $$ \widehat{\mathrm{Adv}}^{\mathsf{pair}}_{T,S} = \left| \widehat{\mathrm{Acc}}^{\mathsf{pair}}_{T,S} - \frac{1}{2} \right|. $$
Thus, distillation makes teacher identification harder under this pairwise adversary, while the remaining advantage shows that some residual teacher-like cues are still detectable.

\subsection{Residual Distinguishability Is Localized}
\label{subsec:discussion-category}

The category-wise results show that residual distinguishability is not uniform across the prompt suite. Aggregate discriminator results can suggest that a student remains distinguishable, but category-wise analyses reveal where that distinguishability is concentrated. For Qwen LoRA, pairwise residual advantage is most visible in style/format, robustness perturbation, domain-technical, and coding prompts. These categories likely expose artifacts related to formatting conventions, precision of technical explanations, handling of noisy instructions, and model-specific response structure. In contrast, safety-boundary, ambiguous, general QA, and summarization prompts are closer to chance under the pairwise judge. This supports a category-aware interpretation: a distilled model may be difficult to distinguish in some behavioral regions while remaining separable in others.

\subsection{Policy-Level Behavior Requires a Separate Layer}
\label{subsec:discussion-policy}

Policy-level behavior should not be inferred solely from semantic similarity or overall discriminator performance. A student may preserve the broad meaning of a teacher response but weaken caveats, omit professional guidance, or change whether it refuses a sensitive request. Conversely, it may use different wording while preserving the same policy-level decision. Our framework captures this distinction through a policy evaluator $\pi(x,y)$, which maps prompt-response pairs to policy-relevant outcomes. This supports the notion of policy-level agreement or disagreement (refer Section~\ref{subsec:policy-indistinguishability} and Equation~\eqref{eq:bound-indistinguishinility:policy-disagreement}).

\subsection{Coverage Matters for Query Acquisition}
\label{subsec:discussion-query}

The query-budget experiments show that more teacher queries generally improve semantic similarity, but disagreement-guided acquisition does not consistently outperform stratified random sampling. This suggests that high-disagreement examples are not automatically the most useful examples for behavioral distillation. A plausible explanation is that global disagreement selection can over-focus on outliers or high-variance prompts while reducing coverage of the broader behavioral space. Category-balanced disagreement partially addresses this, but the results indicate that coverage and diversity are strong baselines. For future acquisition strategies, the objective should not be disagreement alone, but disagreement combined with category coverage, diversity, and possibly discriminator-in-the-loop prompt selection.

\subsection{Implications for Black-Box Distillation}
\label{subsec:discussion-implications}

The main implication is that black-box LLM distillation is better evaluated as a bounded adversarial measurement problem rather than through semantic similarity alone. Semantic similarity measures whether the student captures the teacher's content, while distinguishing advantage measures whether residual artifacts remain detectable. Policy-level and category-wise evaluations further show whether safety-relevant behavior transfers and where imitation succeeds or fails.

Together, these measurements provide a more informative view than any single metric. Our results show that LoRA distillation reduces teacher---student distinguishability under several empirical adversaries, but does not eliminate all detectable differences. This is the intended role of bounded behavioral indistinguishability: not to claim that two models are identical, but to specify how difficult they are to distinguish under explicit query, computational, and adversary-class constraints.

\section{Limitations}
\label{sec:limitations}

The purpose of the paper is not to claim universal indistinguishability between teacher and student models. The results should be interpreted as bounded empirical evidence under the prompt suite, model families, query budgets, decoding settings, and adversary classes studied in this paper. Following are the limitations:

\begin{itemize}
    \item \textbf{Prompt distribution: } The behavioral probe suite is controlled and metadata driven rather than drawn from natural deployment traffic. This design is useful because it lets us evaluate distinguishability across targeted behavioral regions such as reasoning, coding, ambiguity, safety-boundary prompts, instruction conflict, and style control. However, the measured advantages are specific to this probe distribution. 
    
    \item \textbf{Adversary coverage:} We instantiate several empirical adversaries, including learned discriminators, embedding-based measurements, policy-level metrics, and a pairwise automated judge. However, these adversaries are not exhaustive. Stronger adaptive discriminators, human expert judges, or discriminator-in-the-loop prompt generation could reveal additional residual distinguishability.


    \item \textbf{Model and distillation scope: } We evaluate two open teacher--student families, Qwen and Llama, under a practical parameter-efficient LoRA distillation setting. 
    The same framework can naturally be extended to larger models, cross-family distillation, full fine-tuning, preference optimization, or multi-stage distillation to study how different training regimes affect residual distinguishability.


    \item \textbf{Statistical and adaptive evaluation: } Empirical advantages are estimated from finite held-out prompt sets, with category-wise results reflecting smaller per-category sample sizes. Future extensions can add confidence intervals, larger held-out sets, and adaptive adversaries that select prompts based on previous teacher--student responses.
\end{itemize}
Overall, these points clarify the intended scope of bounded behavioral indistinguishability. 

\section{Conclusion}
\label{sec:conclusion}

We introduced bounded behavioral indistinguishability as a security-inspired framework for evaluating whether black-box distillation makes a student merely similar to its teacher, or difficult to distinguish from it under explicit empirical constraints. Central to the framework is $(\epsilon,q,t,\mathbb{A})$-behavioral indistinguishability, which bounds the distinguishing advantage of an adversary class $\mathbb{A}$ under a query budget $(q)$ and computational budget $(t)$. This formulation provides a reproducible way to measure residual behavioral distinguishability between a teacher and a student model.

Across Qwen and Llama model families, LoRA distillation improves semantic similarity between teacher and student responses and reduces learned-discriminator separability. A pairwise teacher-identification experiment using a different-family Llama judge further shows that Qwen pairwise distinguishing advantage decreases from $(0.158)$ for the base student to $(0.081)$ after LoRA distillation.  Our results also show that residual distinguishability is heterogeneous across behavioral categories and adversary classes. Style/format, robustness, and domain-technical prompts remain more distinguishable than some other categories, and naive disagreement-guided query acquisition does not consistently outperform stratified random sampling. These findings suggest that semantic similarity is useful but insufficient. Bounded behavioral indistinguishability shifts the evaluation question from ``How similar are the outputs?" to ``Under explicit query, computational, and adversary-class constraints, can an adversary tell the teacher and student apart?"

\bibliographystyle{unsrt}
\bibliography{references}

\end{document}